\documentclass[10pt,twocolumn,letterpaper]{article}
\usepackage[pagenumbers]{cvpr} 

\usepackage{graphicx}
\usepackage{amsmath}
\usepackage{amssymb}
\usepackage{booktabs}

\usepackage{multirow}
\usepackage{siunitx}
\usepackage{pifont}
\usepackage{makecell}
\usepackage[dvipsnames]{xcolor}

\let\OLDthebibliography\thebibliography
\renewcommand\thebibliography[1]{
  \OLDthebibliography{#1}
  \setlength{\parskip}{1.33pt}
  \setlength{\itemsep}{8pt plus 0.1ex}
}

\usepackage[pagebackref,breaklinks,colorlinks]{hyperref}

\usepackage[capitalize]{cleveref}
\crefname{section}{Sec.}{Secs.}
\Crefname{section}{Section}{Sections}
\Crefname{table}{Table}{Tables}
\crefname{table}{Tab.}{Tabs.}

\usepackage[accsupp]{axessibility}  

\makeatletter
\def\blfootnote{\gdef\@thefnmark{}\@footnotetext}
\makeatother
\def\cvprPaperID{10087} 
\def\confName{CVPR}
\def\confYear{2023}
\begin{document}

\title{SynthVSR: Scaling Up Visual Speech Recognition With Synthetic Supervision}

\author{Xubo Liu$^{1}\textsuperscript{$\ast$}$, Egor Lakomkin$^2$, Konstantinos Vougioukas$^2$, Pingchuan Ma$^2$, Honglie Chen$^2$, Ruiming Xie$^2$, \\Morrie Doulaty$^2$, Niko Moritz$^2$, J\'{a}chym Kol\'{a}\v{r}$^2$, Stavros Petridis$^2$, Maja Pantic$^2$, Christian Fuegen$^2$
\and 
$^1$University of Surrey
\and 
$^2$Meta AI
}

\twocolumn[{%
\renewcommand\twocolumn[1][]{#1}%
\maketitle
\begin{center}
    \vspace{-15pt}
    \includegraphics[width=\linewidth]{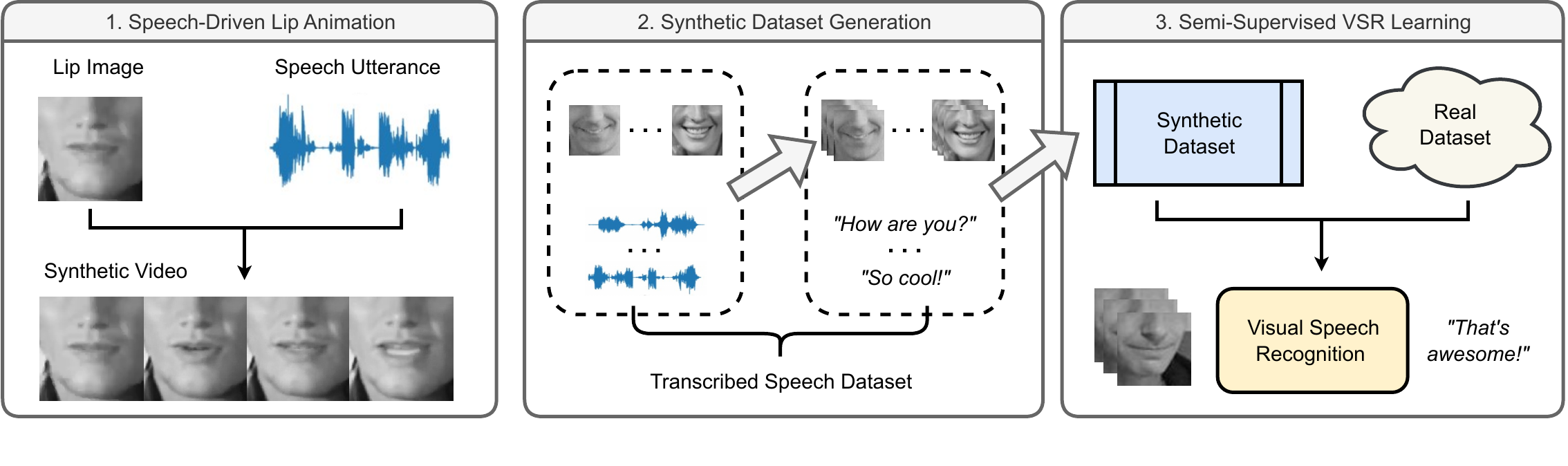}
    \vspace{-2em}
    \captionof{figure}{\textbf{Scaling up visual speech recognition with synthetic supervision (SynthVSR):} we propose SynthVSR, a semi-supervised framework that can substantially improve the performance of VSR models by using synthetic lip movements. Firstly, we introduce a speech-driven lip animation model that generates lip movement videos conditioned on input lip images and speech utterances (left). Secondly,  we generate large-scale synthetic videos using transcribed speech datasets and lip images. The combination of synthetic videos and their corresponding speech transcriptions constitutes the synthetic dataset (centre). Finally, we conduct semi-supervised VSR training with synthetic and real datasets. Our method substantially improves the performance of VSR models with large-scale synthetic data (right).} 
    \label{fig:cvpr}
\end{center}%
}]

\begin{abstract}
\blfootnote{\textsuperscript{$\ast$}Work done during an internship at Meta AI.}

Recently reported state-of-the-art results in visual speech recognition (VSR) often rely on increasingly large amounts of video data, while the publicly available transcribed video datasets are limited in size. In this paper, for the first time, we study the potential of leveraging synthetic visual data for VSR. Our method, termed SynthVSR, substantially improves the performance of VSR systems with synthetic lip movements. The key idea behind SynthVSR is to leverage a speech-driven lip animation model that generates lip movements conditioned on the input speech. The speech-driven lip animation model is trained on an unlabeled audio-visual dataset and could be further optimized towards a pre-trained VSR model when labeled videos are available. As plenty of transcribed acoustic data and face images are available, we are able to generate large-scale synthetic data using the proposed lip animation model for semi-supervised VSR training. We evaluate the performance of our approach on the largest public VSR benchmark - Lip Reading Sentences 3 (LRS3). SynthVSR achieves a WER of 43.3\% with only 30 hours of real labeled data, outperforming off-the-shelf approaches using thousands of hours of video. The WER is further reduced to 27.9\% when using all 438 hours of labeled data from LRS3, which is on par with the state-of-the-art self-supervised AV-HuBERT method. Furthermore, when combined with large-scale pseudo-labeled audio-visual data SynthVSR yields a new state-of-the-art VSR WER of 16.9\% using publicly available data only, surpassing the recent state-of-the-art approaches trained with 29 times more non-public machine-transcribed video data (90,000 hours). Finally, we perform extensive ablation studies to understand the effect of each component in our proposed method.
\end{abstract}

\vspace{-1em}
\section{Introduction}
\label{sec:intro}
Visual speech recognition (VSR), also known as lip reading, is the task of recognizing speech content based on visual lip movements. VSR has a wide range of applications in real-world scenarios such as helping the hearing-impaired perceive human speech and improving automatic speech recognition (ASR) in noisy environments. 

VSR is a challenging task, as it requires capturing speech from high-dimensional spatio-temporal videos, while multiple words are visually ambiguous (e.g., ``world" and ``word") in the visual streams. 
Recently, with the release of large-scale transcribed audio-visual datasets such as LRS2 \cite{afouras2018deep} and LRS3 \cite{afouras2018lrs3}, deep neural networks have become the mainstream approach for VSR. However, even the largest public dataset for English VSR, LRS3, does not exceed 500 hours of transcribed video data. The lack of large-scale transcribed audio-visual datasets potentially results in VSR models which could only work in a laboratory environment i.e. limited word vocabulary and lip sources diversity \cite{ma2022nature}.

A common solution to this issue is to collect and annotate large-scale audio-visual datasets. For example, \cite{serdyuk2022transformer, serdyuk2021audio} collected 90,000 hours of YouTube videos with user-uploaded transcriptions to achieve state-of-the-art performance on standard benchmarks. However, such a procedure is expensive and time-consuming, especially for most of the world's 7,000 languages \cite{shi2022learning}. If annotations are missing, the ASR can be used to generate the transcriptions automatically and this has been shown to be an effective approach to significantly improve VSR performance \cite{ma2022nature}. The other promising direction is to learn audio-visual speech representations from large amounts of parallel unlabeled audio-visual data in a self-supervised approach, and then fine-tune them on the limited labeled video dataset \cite{shi2022learning}. Nevertheless, publicly available video datasets are also limited and their usage may raise license-related\footnote{Such as LRW \cite{chung2016lip} and LRS2 \cite{afouras2018deep} datasets which are only permitted for the purpose of academic research.} concerns, barring their use in commercial applications.

Human perception of speech is inherently multimodal, involving both audition and vision \cite{shi2022learning}. ASR, which is a complementary task to VSR, has achieved impressive performance in recent years, with tens of thousands of hours of annotated speech datasets \cite{panayotov2015librispeech, hernandez2018ted, ardila2019common} available for large-scale training. It is intuitive to ask: \textit{Can we improve VSR with large amounts of transcribed acoustic-only ASR training data?} The key to this question is to take advantage of recent advances in speech-driven visual generative models \cite{vougioukas2018end, vougioukas2020realistic}. By leveraging visual generative models, we can produce parallel synthetic videos for large-scale labeled audio datasets. Synthetic videos provide advantages such as having control over the target text and lip image as well as the duration of a generated utterance. To the best of our knowledge, the potential of leveraging synthetic visual data for improving VSR has never been studied in the literature.

In this work, we present SynthVSR, a novel semi-supervised framework for VSR. In particular, we first propose a speech-driven lip animation model that can generate synthetic lip movements video conditioned on the speech content. Next, the proposed lip animation model is used to generate synthetic video clips from transcribed speech datasets (e.g., Librispeech \cite{panayotov2015librispeech}) and human face datasets (e.g., CelebA \cite{liu2015faceattributes}). Then, the synthetic videos together with the corresponding transcriptions are used in combination with the real video-text pairs (e.g., LRS3 \cite{afouras2018lrs3}) for large-scale semi-supervised VSR training. The pipeline of SynthVSR is illustrated in \cref{fig:cvpr}. Unlike existing studies in exploiting unlabeled video data for VSR using methods such as pseudo-labeling \cite{ma2022nature} and self-supervised learning \cite{shi2022learning}, we use the unlabeled audio-visual data to train a cross-modal generative model in order to bridge ASR training data and VSR. Furthermore, we propose to optimize the lip animation model towards a pre-trained VSR model when labeled videos are available. We empirically demonstrate that the semantically high level, spatio-temporal supervision signal from the pre-trained VSR model offers the lip animation model more accurate lip movements.

SynthVSR achieves remarkable performance gains with labeled video data of different scales. We evaluate the performance of SynthVSR on the largest public VSR benchmark LRS3 with a Conformer-Transformer encoder-decoder VSR model \cite{ma2022nature}. In the low-resource setup using only 30 hours of labeled video data from LRS3 \cite{afouras2018lrs3}, our approach achieves a VSR WER of 43.3\%, substantially outperforming the former VSR methods using hundreds or thousands of hours of video data for supervised training \cite{ren2021learning, afouras2018deep, xu2020discriminative, shillingford2018large, ma2021end} and self-supervised learning \cite{shi2022learning, ma2021lira, afouras2020asr}. Notably, we demonstrate the first successful attempt that trains a VSR model with considerable WER performance using only 30 hours of real video data. Using the complete 438 hours from LRS3 further improves WER to 27.9\% which is on par with the state-of-the-art self-supervised method AV-HuBERT-LARGE \cite{shi2022learning} that uses external 1,759 hours of unlabeled audio-visual data, but with fewer model parameters.
Furthermore, following a recent high-resource setup \cite{ma2022auto} which uses additional 2,630 hours of ASR pseudo-labeled publicly available audio-visual data, our proposed method yields a new state-of-the-art VSR WER of 16.9\%, surpassing the former state-of-the-art approaches \cite{serdyuk2022transformer, serdyuk2021audio} trained on 90,000 hours of non-public machine-transcribed data.

Finally, we present extensive ablation studies to analyze where the improvement of SynthVSR comes from (e.g., the diversity of lip sources, the scale of ASR data). We also show considerable VSR improvement using synthetic video data derived from Text-To-Speech (TTS)-generated speech, indicating the great potential of our method for VSR.

\section{Related Work}
\label{sec:intro}
\noindent\textbf{Visual Speech Recognition.}
VSR has achieved remarkable progress with the success of deep learning and the availability of audio-visual datasets such as LRS2 \cite{afouras2018deep} and LRS3 \cite{afouras2018lrs3}. Progress was driven by adapting ASR approaches such as sequence-to-sequence models \cite{chung2016lip} and the Connectionist Temporal Classification (CTC) \cite{assael2016lipnet, graves2006connectionist} training objective. Recent studies have achieved noticeable results in VSR by using Transformer-based architecture \cite{afouras2018deep}, a convolutional variant \cite{zhang2019spatio}, Conformer-based models \cite{gulati2020conformer}, and an attention-based feature pooling method \cite{prajwal2022sub}. The recent advances in VSR models are mainly dependent on leveraging increasingly large-scale transcribed non-public video datasets such as LSVSR (3,800 hours) \cite{shillingford2018large} and YT31k (31,000 hours) \cite{ma2022nature}. Some recent works \cite{serdyuk2021audio, serdyuk2022transformer} use 90,000 hours of non-publicly machine-transcribed video data to achieve state-of-the-art performance on standard benchmarks. Another popular trend is to leverage unlabeled audio-visual datasets such as AVSpeech \cite{ephrat2018looking} and VoxCeleb2 \cite{chung2018voxceleb2} with semi- or self-supervised learning. Using pre-trained ASR models to transcribe audio streams in unlabeled audio-visual datasets i.e., pseudo-labeling has demonstrated great potential to improve VSR performance. Such pseudo-labeled video data can be used for supervised training \cite{ma2022nature} or knowledge distillation \cite{afouras2020asr}. Self-supervised approaches such as contrastive learning \cite{pan2022leveraging, ma2021contrastive, chung2020seeing} and cross-modal feature prediction \cite{ma2021lira} have achieved significant improvement by exploiting unlabeled audio-visual data. AV-HuBERT \cite{shi2022learning} is the state-of-the-art self-supervised method that predicts iteratively refined cluster assignments
from masked audio-visual streams and achieves impressive VSR performance, especially when only limited labeled video data is available (e.g., 30 hours). 

\noindent\textbf{Speech-Driven Facial Animation.} Speech-driven facial animation aims to produce a talking face video conditioned on speech content.
With the advances in deep visual generative models, speech-driven facial animation has significantly improved over the past years. Speech2Vid \cite{chung2017you} is the first end-to-end neural network-based method for speech-driven facial animation. This method is optimized towards a pixel-level $L_1$ reconstruction loss which results in blurry generated frames. Recently, Generative Adversarial Networks (GANs) have made impressive progress \cite{vougioukas2018end, vougioukas2020realistic} in facial animation, as they are able to generate sharp and highly textured facial frames. Initially, most of these methods synthesized frontal faces and focused on improving lip-sync \cite{chen2018lip, vougioukas2020realistic}, preserving the identity of the target speaker \cite{zhou2019talking}. As the generation of synthetic frontal faces improved, researchers began to focus on generating talking heads that could handle natural head movements\cite{zhou2021pose, chen2020talking}.

\noindent\textbf{Learning from Synthetic Supervision.}
Training with synthetic data has attracted increasing research attention in recent years. Synthetic data could offer improved data diversity \cite{doersch2019sim2real}, which is useful in many tasks where labeled data is limited, such as image classification \cite{gan2020threedworld}, objection detection \cite{peng2015learning, prakash2019structured}, semantic segmentation \cite{ros2016synthia, wang2020differential}, human pose estimation \cite{doersch2019sim2real}, and ASR \cite{fazel2021synthasr, mavandadi2021deliberation}. Unlike previous work, we focus on a new multimodal problem: how to produce and leverage synthetic video data for improving VSR. 

\section{SynthVSR}
In  this section, we introduce SynthVSR, a novel semi-supervised VSR framework to improve the performance of VSR models using synthetic data, as shown in \cref{fig:cvpr}. We propose a speech-driven lip animation model trained on an audio-visual dataset, which is used to produce synthetic videos with large-scale transcribed speech and face datasets for scaling up VSR training. We will introduce the VSR model, the speech-driven lip animation model, and the semi-supervised VSR training next.

\begin{figure*}[t]
  \centering
  \includegraphics[width=\linewidth]{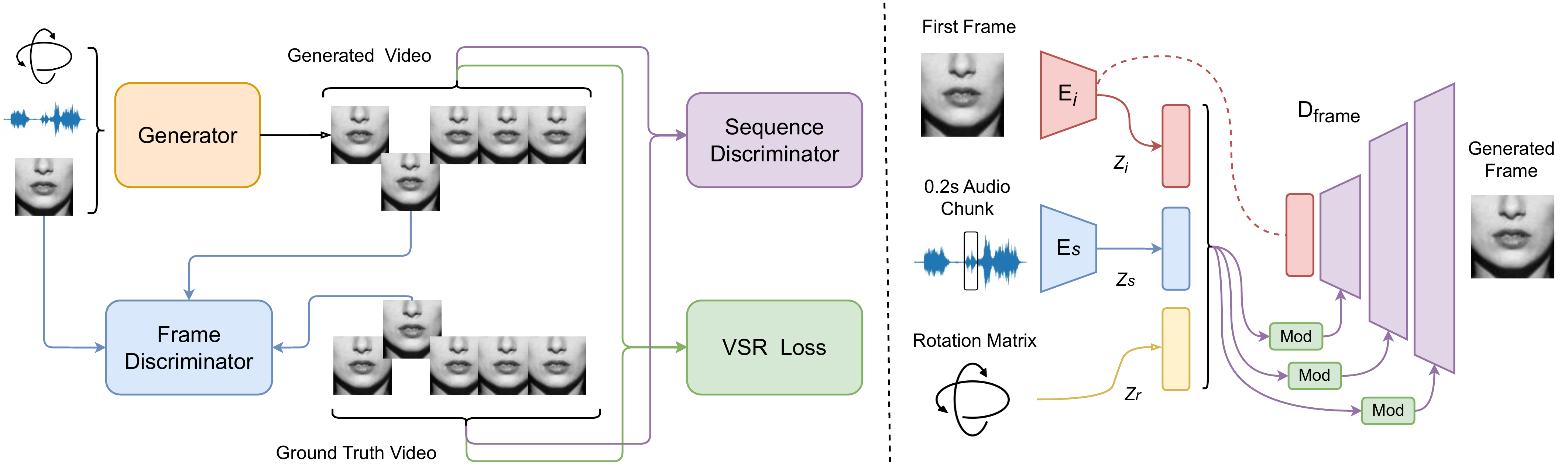}
  \caption{\textbf{Architecture of proposed speech-driven lip animation model.} Left: GAN-based speech-driven lip animation model generating lip movements given a lip image, a speech utterance, and a rotation sequence; Right: structure of the generator in the lip animation model.}
  \vspace{-1.5em}
  \label{fig:lam}
\end{figure*}
\subsection{VSR Model}
\label{sec:vsr-baseline}
The VSR model we use in this work is based on \cite{ma2022nature, ma2021end}, which has achieved the state-of-the-art VSR performance on LRS3 \cite{afouras2018lrs3} without the use of external data. The baseline VSR model is an encoder-decoder architecture. In particular, the encoder is comprised of two components, the visual front-end (a 3D convolutional layer followed by a ResNet-18 model \cite{he2016deep, stafylakis2017combining}) and a Conformer \cite{gulati2020conformer} encoder. The decoder is based on the transformer architecture \cite{vaswani2017attention}. The baseline VSR model is trained end-to-end using a combination of the CTC loss \cite{assael2016lipnet, shillingford2018large} with an attention-based Cross-Entropy (CE) loss. Model details are described in the supplementary material.

\subsection{Speech-Driven Lip Animation}
Inspired by the recent advances in speech-driven facial animation \cite{zhou2019talking, zhou2021pose, vougioukas2018end, vougioukas2020realistic}, we propose an approach for speech-driven lip animation that generates videos of talking mouth regions conditioned on speech utterances. The output space of the lip animation model is the same as the VSR input space. The proposed lip animation model is based on a temporal GAN \cite{vougioukas2018end, vougioukas2020realistic} with two discriminators. We further propose a VSR perceptual loss when labeled video data is available. The architecture of the speech-driven lip animation model is illustrated in the left part of \cref{fig:lam}. We will introduce each component in the next sections.

\noindent\textbf{Generator.}
The generator $G$ is an encoder-decoder structure, as shown in the right part of the \cref{fig:lam}. The generator uses the first frame of a video clip, a speech clip, and a head rotation sequence as inputs. The head rotation is used as the additional condition which helps better model the lip movements as most training videos are not static. The speech clip is divided into overlapping 200 ms chunks with a stride of 40 ms. The generator produces the corresponding video frame for each speech chunk. In the generator, an image encoder $E_i$ and a speech encoder $E_s$ are used to capture the visual information and speech context into latent embeddings $z_i$ and $z_s$, respectively. The image encoder $E_i$ uses a stack of 2D convolutional layers to extract the visual embedding. The speech encoder $E_s$ comprises a stack of 1D convolutional layers followed by a stack of GRU layers. The head rotations are provided in the form of sequences of 3D rotation matrices \cite{zhou2021pose} $z_r \in \mathbb{R}^{3\times3}$ with respect to the first frame. The three embeddings $z_i$, $z_s$ and $z_r$ are concatenated and used to modulate the convolutional layers in the frame decoder $D_{frame}$, which is similar to StyleGAN2 \cite{viazovetskyi2020stylegan2}. Different from StyleGAN2, which generates frames from a learned constant input, we use the penultimate layer features of the image encoder $E_i$ as the input. 

\noindent\textbf{Discriminators.}
The speech-driven lip animation system has two discriminators: frame discriminator $D_{img}$ and sequence discriminator $D_{seq}$. The frame discriminator is a stack of CNN layers that operates on the image frame level. The original first frame is concatenated channel-wise to the target frame to form the input of the frame discriminator. This helps enforce visual consistency. The sequence discriminator operates on the sequence level to ensure the temporal consistency of synthetic lip movements. The sequence discriminator uses spatio-temporal convolutions to encode the image sequence, followed by GRU layers to determine if the sequence is real or not. Specifically, the frame discriminator $D_{img}$ is trained on frames that are uniformly sampled from a video $v$ using a sampling function $S(v)$. The first frame $v_1$ is fed to the $D_{img}$ as the condition. The input speech signal is $s$. The adversarial loss of the $D_{img}$ is defined as:
\begin{equation}
\begin{split}
\label{eq-1}
\mathcal{L}_{Disc}^{img} = \mathbb{E}_{v}[\operatorname{log}D_{img}(S(v),v_1)] \\ + \mathbb{E}_{v,s}[\operatorname{log}(1-D_{img}(S(G(s, v_1)), v_1)].
\end{split}
\end{equation}
$D_{seq}$ operates on the entire sequence video $v$. The adversarial loss of the $D_{seq}$ is defined as follows:
\begin{equation}
\begin{split}
\label{eq-1}
\mathcal{L}_{Disc}^{seq} = \mathbb{E}_{v}[\operatorname{log}D_{seq}(v)] + \\ \mathbb{E}_{v,s}[\operatorname{log}(1-D_{seq}(G(s, v_1))].
\end{split}
\end{equation}

\noindent\textbf{VSR Perceptual Loss.}
We further propose to optimize the lip animation model towards a VSR perceptual loss if labeled video data is available. We first pre-train a VSR model as introduced in \cref{sec:vsr-baseline}. The proposed VSR perceptual loss corresponds to a weighted sum of feature distances computed from the visual front-end and the Transformer decoder of the pre-trained VSR model for real and generated samples. We use $L_1$ norm to measure the visual embedding distance and Kullback–Leibler (KL) divergence to measure the logits distribution distance, respectively. The VSR perceptual loss is obtained by:
\vspace{-0.5em}
\begin{equation}
\begin{split}
    \label{eq-vsr}
    \mathcal{L}_{VSR} = \lambda_{visual}\left\lVert z_f^{r} - z_f^{s} \right\lVert_1 + \lambda_{logits}\operatorname{KL}(\hat y^{r}, \hat y^{s}),
\end{split}
\end{equation}
where $z_f^{r}$ and $z_f^{s}$ are the VSR front-end visual features of the real and synthetic video, respectively, $\hat y^{r}$ and $\hat y^{s}$ is the VSR predicted logits distribution of real and synthetic video, respectively, $\lambda_{visual}$ and $\lambda_{logits}$ control the weights of these two perceptual losses. The VSR model is frozen during the lip animation model training.

\noindent\textbf{Training Objectives.}
The speech-driven lip animation model is trained using a combination of a reconstruction loss, adversarial losses, and a VSR perceptual loss. The reconstruction loss is computed based on the $L_1$ distance between the generated video $\hat v$ and ground truth video $v$: 
\begin{equation}
    \label{eq-1}
    \mathcal{L}_{rec} = \left\lVert v -\hat v \right\lVert_1.
\end{equation}
The overall training loss for the lip animation model is:
\begin{equation}
\begin{split}
\label{eq-lam}
\mathcal{L}_{Animation} =  
\lambda_{disc}^{img}\mathcal{L}_{disc}^{img}
 + \ \lambda_{disc}^{seq}\mathcal{L}_{disc}^{seq} \\ + \lambda_{rec}\mathcal{L}_{rec} + \mathcal{L}_{VSR},
\end{split}
\end{equation}
where $\lambda_{disc}^{img}$, $\lambda_{disc}^{seq}$, and $\lambda_{rec}$ represent the coefficient of the adversarial loss of frame discriminator $D_{img}$, the adversarial loss of sequence discriminator $D_{seq}$, and the reconstruction loss, respectively.

\subsection{Semi-Supervised VSR with Synthetic Data}
In a typical supervised setting, a VSR model is trained from the labeled dataset $\mathcal{D}_{real}={\{(v_i^r,y_i^r)\}_{i=1}^{n_r}}$, where $n_r$ is the number of paired video clips $v_i^r$ and its transcriptions $y_i^r$. In SynthVSR, we first train a speech-driven lip animation model on an unlabeled audio-visual dataset $\mathcal{D}_{av}={\{(v_i^{av},s_i^{av})\}_{i=1}^{n_{av}}}$, where $n_{av}$ is the number of paired video clips $v_i^{av}$ and speech utterances $s_i^{av}$. Furthermore, we can use the trained generator $G$ to generate synthetic video from a labeled speech dataset $\mathcal{D}_s={\{(s_i^s,y_i^s)\}_{i=1}^{n_s}}$ and a cropped lips dataset $\mathcal{D}_f={\{c_i^f\}_{i=1}^{n_f}}$, where $n_s$ is the number of paired speech clips $s_i^s$ and its transcriptions $y_i^s$ and $n_f$ is the number of lip images $c_i^f$. For each speech clip $s_i^s$, the generator $G$ generates its parallel synthetic video $\hat {v}_i^s$:
\begin{equation}
    \label{eq-2}
    G(s_i^s, S(\mathcal{D}_f)) \mapsto \hat {v}_i^s,
\end{equation}
where $S(\cdot)$ is a uniform sampling function. Then, the synthetic dataset $\mathcal{D}_{synth}={\{\hat {v}_i^s,y_i^s)\}_{i=1}^{n_s}}$ is obtained, where in general $n_s >> n_r$ as there are far more annotated speech datasets (e.g., Librispeech \cite{panayotov2015librispeech}) than labeled video datasets. Finally, the VSR model is trained from $\mathcal{D}_{real} \cup \mathcal{D}_{synth}$, where the rich visual and text label information in $\mathcal{D}_{synth}$ offers significant performance improvement for VSR.

\section{Experiments}
\begin{table*}[t]
\resizebox{\textwidth}{!}{
\begin{tabular}{ccccccc}
\Xhline{2\arrayrulewidth}
Method    
& Backbone & LM                   & Labeled data (hrs) & Unlabeled data (hrs) & Synthetic data (hrs) & WER (\%) \\ \hline
Afouras et al. \cite{afouras2020asr}  & CNN & \ding{51}  & $\text{595}^\ddagger$                & 334                 & -                    & 59.8 \\
Ren et al.\cite{ren2021learning} & Transformer & \ding{55}                & $\text{818}^\ddagger$                & -                 & -                    & 59.0 \\
Afouras et al. \cite{afouras2018deep}  & Transformer & \ding{51}  & $\text{1,519}^{\dagger\ddagger}$                & -                 & -                    & 58.9 \\
Xu et al. \cite{xu2020discriminative} & RNN & \ding{55} & $\text{595}^\ddagger$                & -                 & -                    & 57.8 \\
Shillingford et al. \cite{shillingford2018large} & RNN & \ding{51}                & $\text{3,886}^\dagger$                & -                 & -                    & 55.1     \\
Ma et al. \cite{ma2021lira} & Transformer  & \ding{55}                & 433                & 1,759                 & -                    & $\text{49.6}^\ast$     \\
Ma et al. \cite{ma2021end} & Conformer  & \ding{51}                & 438                & -                 & -                    & 46.9     \\
AV-HuBERT-BASE \cite{shi2022learning}                 & Transformer & \ding{55}                & 30                & 1,759                 & -                    & 46.1     \\
\hline\multirow{4}{*}{SynthVSR}     & \multirow{4}{*}{Conformer-BASE} &  \ding{55} & 30               & -                    & -                    & 104.0     \\
                                   &                        & \ding{55}     & -                & -                  & 3,652                 & 100.3      \\
                                   &                        & \ding{55}     & 30                & -                  & 3,652                 & 44.7      \\
                                   &                        & \ding{51}     & 30                & -                  & 3,652                 & \textbf{43.3}     \\
\Xhline{2\arrayrulewidth}
\end{tabular}}
\caption{Experimental results of low-resource labeled data setting on LRS3 (test). LM denotes whether or not a language model is used in the decoding.
$^\dagger$Includes non-publicly available data. $^\ddagger$Includes datasets that are only permitted for the purpose of academic research.  hrs is an abbreviation for hours.  $^\ast$Result taken from \cite{shi2022learning}.}
\label{tab:1}
\vspace{-1em}
\end{table*}

\subsection{Datasets}
We use several public datasets in this work. (1) Audio-visual datasets: LRS3 \cite{afouras2018lrs3}, AVSpeech \cite{ephrat2018looking} and VoxCeleb2 \cite{chung2018voxceleb2}; (2) Speech datasets: Librispeech \cite{panayotov2015librispeech}, TED-LIUM 3 \cite{hernandez2018ted}, Common Voice \cite{ardila2019common}; (3) Facial dataset: CelebA \cite{liu2015faceattributes}. 

\noindent\textbf{LRS3.} We conduct experiments on the LRS3 \cite{afouras2018lrs3} dataset, which is the largest public benchmark for English VSR containing  438.9 hours of video clips from TED talks (408, 30, and 0.9 hours in the pre-training, training-validation, and test set, respectively).

\noindent\textbf{AVSpeech \& VoxCeleb2.}
A recent work \cite{ma2022auto} uses public language classifier \cite{valk2021voxlingua107} and ASR models \cite{baevski2020wav2vec} to filter for English content and obtain pseudo-transcriptions for two multilingual audio-visual datasets: AVSpeech \cite{ephrat2018looking} and VoxCeleb2 \cite{chung2018voxceleb2}, containing 1,323 hours from AVSpeech and 1,307 hours from VoxCeleb2. We follow the same data setting for our high-resource setting. Furthermore, we filter out AVSpeech videos with large jitter resulting in 933 hours of videos overall for speech-driven lip animation training.

\noindent\textbf{Datasets for Speech-Driven Lip Animation Training.}
The lip animation model is trained on a combination of LRS3 (pre-training and training-validation splits) and the English subset (933 hours) of AVSpeech datasets.

\noindent\textbf{Datasets for Synthetic Data Generation.}
We use Librispeech \cite{panayotov2015librispeech}, TED-LIUM 3 \cite{hernandez2018ted}, Common Voice (English split) \cite{ardila2019common} datasets as the speech sources. We split Librispeech audio clips into segments of less than 6 seconds based on the provided silence annotations. For TED-LIUM 3 and Common Voice, we filter out the audio clips which are longer than 20s. As a result, 944 hours, 465 hours, and 2,243 hours of speech data are obtained for Librispeech, TED-LIUM 3, and Common Voice, respectively. We use the CelebA \cite{liu2015faceattributes} dataset as a source of lip images, which has 202,599 face images and 10,177 identities. For each speech clip, we randomly sample one image from CelebA to generate one synthetic video. We use a static rotation matrix in the generation process. In total, 3,652 hours of synthetic video clips are generated for scaling up VSR training.

\subsection{Data Processing}
We train a mouth detection module using the Faster R-CNN object detection architecture \cite{ren2015faster} on the CelebA \cite{liu2015faceattributes} dataset. We leverage the mouth corner positions provided in the dataset to identify the area around the mouth in each video frame. With this network, we locate and extract a $96 \times 96$ bounding box around the mouth, which we subsequently convert to grayscale. We use the same video pre-processing methods for lip animation and VSR models. For the text vocabulary, we use SentencePiece \cite{kudo2018subword} subword units with a vocabulary size of 5,000.

\subsection{Implementation Details}
\noindent\textbf{VSR Model.}
We consider two model configurations: (1) Conformer-BASE (250M) with 12-layer Conformer encoder, 6-layer Transformer decoder, 768 input dimensions, 3,072 feed-forward dimensions, and 16 attention heads; (2) Conformer-LARGE (783M) with 24-layers Conformer encoder, 9-layer Transformer decoder, 1024 input dimensions, 4,096 feed-forward dimensions, and 16 attention heads. For each configuration, the encoder and decoder have the same dimensions and attention heads. 

We use horizontal flipping, random cropping, and adaptive time masking as the VSR data augmentation methods \cite{ma2022nature}. We train the BASE and LARGE models for 75 epochs on 64 A100-GPUs with the AdamW \cite{loshchilov2017decoupled} optimizer, a cosine learning rate scheduler, and a warm-up of 5 epochs. The peak learning rate is \num{1e-3} and \num{8e-4} for the BASE and LARGE models respectively. The number of frames in each batch is limited to 2,400 and 1,600 frames for the BASE and LARGE models, respectively. Following \cite{ma2022nature, ma2022auto}, a pre-trained transformer-based language model is used in the VSR decoding stage. The training configuration is the same as that used in the previous work \cite{ma2022auto}.

\noindent\textbf{Speech-Driven Lip Animation.}
For each training video clip, we sample a sequence of 75 frames that corresponds to 3 seconds. The first frame of the sampled video sequence is used to drive the lip animation. We train the lip animation model for 70 epochs on 32 A100-GPUs with a batch size of 3 per GPU. The details of the module structure in the lip animation model are described in the supplementary material. Adam \cite{kingma2014adam} optimizer is used to train the lip animation model with the learning rate \num{1e-4} for the generator and the frame discriminator. Sequence discriminator uses a smaller learning rate of \num{1e-5}. The weights for each loss term are $\lambda_{disc}^{img}=1$, $\lambda_{disc}^{seq}=0.2$, $\lambda_{rec}=300$ in \cref{eq-lam}.

We consider five lip animation model configurations. The lip animation model without VSR perceptual loss is referred to LAM-Baseline. The lip animation model with the BASE VSR model trained on LRS3 is referred to as LAM-LRS3-VSR-VL, with the $\lambda_{visual}=250$ and $\lambda_{logits}=10$ in \cref{eq-vsr}. Two variants LAM-LRS3-VSR-L and LAM-LRS3-VSR-V are further designed with the $\lambda_{visual}=0$ and $\lambda_{logits}=0$, respectively. These two model configurations are used in ablation studies. Last, the lip animation model with the BASE VSR model trained on LRS3 and 2,630 hours of pseudo-labeled AVSpeech and VoxCeleb2 is referred to as LAM-LRS3-AVoX-VSR, with the $\lambda_{visual}=500$ and $\lambda_{logits}=10$.

\subsection{Low-Resource Labeled Data Setting}
The LAM-Baseline model is used to generate 3,652 hours of synthetic data. We use 30 hours of LRS3 (training-validation) to evaluate the VSR performance when labeled data is scarce. We conduct experiments on the BASE VSR model. The results are reported in \cref{tab:1}.
As supervised VSR training from scratch with long sequences often poses optimization problems \cite{ma2022nature, chung2016lip, afouras2018deep}, we first use 30 hours of LRS3 and the 944 hours of Librispeech synthetic data to pre-train a VSR model with the same architecture as \cite{ma2022nature}, then we take the weights of the pre-trained visual front-end to train the BASE VSR model. The pre-training details are described in the supplementary material. 

Training with synthetic data leads to dramatic performance improvement when only 30 hours of labeled data is available. Using 3,652 hours of synthetic data and 30 hours of LRS3 labeled data for training, the BASE model achieves the WER 43.3\% (44.7\% w./o. language model). Training from 30 hours of LRS3 results in a poor WER of 104\%. Our proposed method significantly outperforms the former approaches using hundreds or thousands of data for supervised training \cite{ren2021learning, afouras2018deep, xu2020discriminative, shillingford2018large, ma2021end} and self-supervised learning \cite{ma2021lira, afouras2020asr}. Furthermore, our method performs better than the AV-HuBERT-BASE \cite{shi2022learning} model (46.1\%) that uses 30 hours of LRS3 and 1,759 hours of unlabeled audio-visual data for self-supervised learning. Notably, we show the first attempt that trains a VSR model with considerable WER performance using only 30 hours of real video data. In addition, we observe that training only from synthetic data performs poorly (100.3\%), which may be caused by the domain mismatch between the real and synthetic data.

\begin{table*}[t]
\resizebox{\textwidth}{!}{
\begin{tabular}{ccccccc}
\Xhline{2\arrayrulewidth}

Method                              & Backbone & LM                   & Labeled data (hrs) & Unlabeled data (hrs) & Synthetic data (hrs) & WER (\%) \\ \hline
AV-HuBERT-BASE \cite{shi2022learning}                 & Transformer & \ding{55}                & 433                & 1,759                 & -                    & 34.8     \\
Makino et al. \cite{makino2019recurrent}                     & Transformer & \ding{55}              & $\text{31,000}^\dagger$         & -                    & -                    & 33.6     \\
Ma et al. \cite{ma2022nature}                         & Conformer  & \ding{51}                 & $\text{1,459}^\ddagger$               & -                    & -                    & 31.5     \\
Prajwal et al. \cite{prajwal2022sub} & Transformer & \ding{51}                & $\text{2,676}^\dagger$                & -                 & -                    & 30.7     \\
AV-HuBERT-LARGE \cite{shi2022learning}  & Transformer & \ding{55}                & 433                & 1,759                 & -                    & 28.6     \\
AV-HuBERT-LARGE w. Self-Training \cite{shi2022learning}  & Transformer & \ding{55}                & 433                & 1,759                 & -                    & 26.9     \\
Auto-AVSR \cite{ma2022auto}       & Conformer    & \ding{51}      & $\text{3,448}^\ddagger$                & -                 & -                    & 19.1     \\
Serdyuk et al. \cite{serdyuk2021audio} & Transformer       & \ding{55}         &   $\text{90,000}^\dagger$              & -                 & -                    & 25.9     \\
Serdyuk et al. \cite{serdyuk2022transformer} & Transformer       & \ding{55}         & $\text{90,000}^\dagger$                & -                 & -                    & 17.0     \\
\hline\multirow{6}{*}{SynthVSR}     & \multirow{6}{*}{Conformer-BASE} &  \ding{55} & 438               & -                    & -                    & 36.7     \\
                                   &                    & \ding{55}         & 438                &  -                & 3,652                  & 28.4        \\
                                   &                    & \ding{51}         & 438                &  -                & 3,652                  & \textbf{27.9}        \\

                                   &                        & \ding{55}     & 3,068                & -                  & -                 & 21.2        \\
                                   &                        & \ding{55}     & 3,068                & -                  & 3,652                 & 19.4       \\
                                   &                        & \ding{51}     & 3,068                & -                  & 3,652                 & \textbf{18.7}    \\
\hline\multirow{2}{*}{SynthVSR}     & \multirow{2}{*}{Conformer-LARGE} &  \ding{55} & 3,068               & -                    & 3,652                    & 18.2     \\
                                  &                    & \ding{51}         & 3,068                &  -                & 3,652                  & \textbf{16.9}        \\

\Xhline{2\arrayrulewidth}

\end{tabular}}
\caption{Experimental results of LRS3 \& high-resource labeled data setting on LRS3 (test). LM denotes whether or not a language model is used in the decoding.
$^\dagger$Includes non-publicly available data. $^\ddagger$Includes datasets that are only permitted for the purpose of academic research. hrs is an abbreviation for hours.}
\label{tab:2}
\vspace{-1em}
\end{table*}
\subsection{LRS3 Labeled Data Setting}
In this setting, we report the results when using the full 438 hours of LRS3. Experiments are conducted on the BASE VSR model. To avoid the optimization problem, we initialize the weights of the visual front-end from a publicly available Conformer-based model \cite{ma2022nature} pre-trained on LRS3 using curriculum learning. The results are shown in \cref{tab:2}.

Our BASE model achieves WER 36.7\% when using 438 hours of LRS3 data for training. We generate 3,652 hours of synthetic data using the LAM-LRS3-VSR-VL model. Using 3,652 hours of synthetic data and 438 hours of LRS3 labeled data, the BASE VSR model achieves the WER 27.9\% (28.4\% w/o language model, corresponding to a WER reduction of 8.3\%). Our method outperforms three recent approaches using 31,000 (33.6\%), 1,459 (31.5\%), and 2,679 (30.7\%) hours of labeled data, respectively. When compared with the state-of-the-art self-supervised method AV-HuBERT \cite{shi2022learning} that uses additional 1,759 hours of unlabeled audio-visual data, our method outperforms the AV-HuBERT-BASE model (34.8\%) by a large margin. Our method slightly performs better than the AV-HuBERT-LARGE model (28.6\%), but with fewer model parameters (our BASE model 250M vs AV-HuBERT-LARGE 390M). Note that we compare with the AV-HuBERT results without self-training as we do not use the pseudo-labeled 933 hours of AVSpeech subset for VSR training. 

\subsection{High-Resource Labeled Data Setting}
We further evaluate the scalability of SynthVSR: when using machine-transcribed AVSpeech \cite{ephrat2018looking} and VoxCeleb2 \cite{chung2018voxceleb2} as additional training data. We generate 3,652 hours of synthetic data with  LAM-LRS3-AVoX-VSR and conduct experiments on BASE and LARGE models. The visual front-end is initialized from a pre-trained model which is the same as that used in the LRS3 labeled data setting.

We first train the BASE model with 438 hours of labeled LRS3 and 2,630 hours of pseudo-labeled data, resulting in a strong VSR baseline with the WER 21.2\%. By using additional 3,652 hours of synthetic data, the WER of the BASE model improves to 18.7\% (19.4\% w./o. language model), which outperforms \cite{ma2022auto} that uses additional labeled dataset LRS2 (223 hours) and LRW (157 hours) for training. Although the VSR model has seen a large amount of labeled data, and the speech-driven lip animation model is trained from part of the VSR training data, synthetic data can still lead to considerable performance gains. Furthermore, increasing the model size from BASE to LARGE results in better VSR performance with the WER of 16.9\% (18.2\% w./o. language model), which is the current state-of-the-art performance on LRS3, with publicly available data only.  

\subsection{Why SynthVSR Improves VSR?}
We conduct extensive ablation studies to understand the impact of VSR perceptual loss, the diversity of lip sources, and the scale of ASR data for SynthVSR. In addition, we assess the domain mismatch between the real and synthetic data using VSR models. We perform experiments on the LRS3 labeled data setting with the Conformer-BASE VSR model.

\noindent\textbf{Effect of VSR Perceptual Loss.}
We experiment with  four lip animation model variants: LAM-Baseline, LAM-LRS3-VSR-V, LAM-LRS3-VSR-L, LAM-LRS3-VSR-VL to generate 944 hours of synthetic data from Librispeech, respectively. The WER of the BASE model trained with only 438 hours of LRS3 data is 36.7\% (see in \cref{tab:2}), the results are shown in \cref{tab:3}. Using 944 hours of synthetic data generated from the LAM-Baseline model, the WER improves to 32.2\%. The individual visual VSR loss and linguistic VSR loss improve the WER to 31.7\% and 31.4\%, respectively. The linguistic VSR loss performs slightly better than the visual one. Further, the combination of visual and linguistic VSR loss further improves the WER to 30.8\%, which indicates that the spatio-temporal visual and high-level semantic knowledge from a pre-trained VSR model could offer the lip animation model more accurate lip movements.
\begin{table}[h]
\centering
\resizebox{0.29\textwidth}{!}{
\begin{tabular}{cc}
\Xhline{2\arrayrulewidth}
\textbf{Lip animation model} & \textbf{WER (\%)}                   \\ \hline
LAM-Baseline  & 32.2 \\
LAM-LRS3-VSR-V  & 31.7 \\
LAM-LRS3-VSR-L & 31.4 \\
LAM-LRS3-VSR-VL  & \textbf{30.8} \\
\Xhline{2\arrayrulewidth}
\end{tabular}}
\caption{Ablation studies on the proposed VSR perceptual loss.}
\label{tab:3}
\end{table}
\vspace{-1em}

\noindent\textbf{Effect of Increasing Lip Sources Diversity.}
To analyze the impact of the diversity of lip sources we generate 944 hours of synthetic data generated from Librispeech using two target face sources: CelebA and LRS3 (pre-training). Images from LRS3 (pre-training) do not introduce any new lip images for VSR training. The results are reported in \cref{tab:4}. Using the synthetic data with LRS3 lip images, the BASE model achieves the WER 32.1\% which is 1.3\% worse than that with CelebA lip images, which means the external lip images from CelebA offer better WER improvement.
\begin{table}[h]
\centering
\resizebox{0.4\textwidth}{!}{
\begin{tabular}{ccc}
\Xhline{2\arrayrulewidth}
\textbf{Lip source} & \textbf{Unique images} & \textbf{WER} (\%)                   \\ \hline
LRS3-pretrain & 5,090 &   32.1 \\
CelebA & 202,599 & \textbf{30.8} \\
\Xhline{2\arrayrulewidth}
\end{tabular}}
\caption{Ablation studies on the impact of lip sources diversity. LRS3-pretrain indicates using the LRS3 (pre-training) images.}
\label{tab:4}
\vspace{-2em}
\end{table}

\noindent\textbf{Effect of the Scale of Speech Corpus.}
Using synthetic data generated from TED-LIUM 3, Librispeech, and Common Voice with LAM-LRS3-VSR-VL model synthetic data, the LRS3 (test) WER improves from 36.7\% to 30.9\%, 30.8\%, and 30.1\%, respectively (see in \cref{tab:5}). When combining all synthetic data together (3,652 hours) for training, the WER improves to 28.4\%. We observe that richer speech corpora can offer larger performance improvements. Furthermore, we find that the improvement from scaling speech corpora is more pronounced than when adding VSR perceptual loss and lip sources diversity, suggesting that our proposed method benefits greatly from large speech datasets.

\begin{table}[h]
\centering
\resizebox{0.45\textwidth}{!}{
\begin{tabular}{ccc}
\Xhline{2\arrayrulewidth}
\textbf{Training data} & \textbf{Hours} & \textbf{WER (\%)}                   \\ \hline 
LRS3  & 438 & 36.7 \\
LRS3 + TED-Synth & 903 & 30.9 \\
LRS3 + LBS-Synth & 1,382 & 30.8 \\
LRS3 + CV-Synth & 2,681 & 30.1 \\
LRS3 + [TED + LBS + CV]-Synth & 4,090  & \textbf{28.4} \\
\Xhline{2\arrayrulewidth}
\end{tabular}}
\caption{Ablation studies on speech corpora. TED-Synth, LBS-Synth, and CV-Synth indicates the synthetic data generated from TED-LIUM 3, Librispeech, and Common Voice, respectively. }
\label{tab:5}
\end{table}

\begin{figure}[h]
  \centering
   \includegraphics[width=1\linewidth]{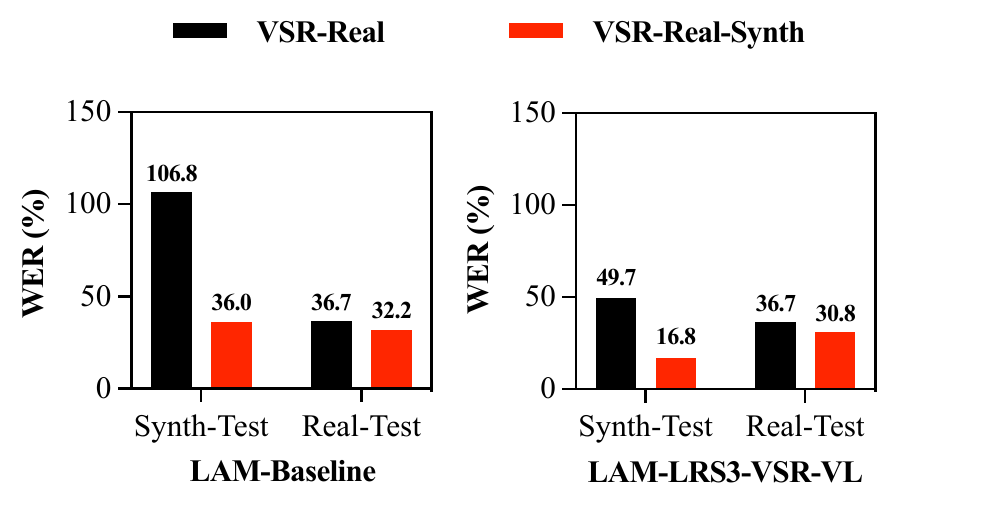}
   \vspace{-2em}
   \caption{Assessment of the domain mismatch between real and synthetic video data using VSR models trained with only real data (VSR-Real, black) and real data together with synthetic data (VSR-Real-Synth, red), respectively. Synth-Test and Real-Test refer to the synthetic and real LRS3 (test) sets.}
   \label{fig:abs}
\end{figure}

\noindent\textbf{Assessment of the Domain Mismatch Using VSR.}
We use 944 hours of Librispeech synthetic data generated from LAM-Baseline and LAM-LRS3-VSR-VL for VSR training. We conduct an assessment on the LRS3 real and synthetic test sets, the synthetic test set is obtained by passing the speech and the first frame in the real test set to the lip animation model. The results are illustrated in \cref{fig:abs}. By introducing the VSR perceptual loss, the WER on the synthetic test set is reduced from 106.8\% to 49.7\%, which indicates the VSR loss minimizes the domain gap between real and synthetic data. After training on synthetic data, the WER reduces for both real and synthetic data. Specifically, the WER is improved substantially from 106.8\% to 36.0\% and from 49.7\% to 16.8\% on synthetic test sets generated using LAM-Baseline and LAM-LRS3-VSR-VL, respectively. We observe that domain mismatch can be further reduced by combining real and synthetic data for VSR training.

\subsection{SynthVSR with TTS-Generated Speech}
In the real world, text corpora (e.g., Wikipedia) are more accessible than transcribed speech corpora. To explore the potential of using TTS-generated speech for SynthVSR, we use the off-the-shelf TTS model FastSpeech 2 \cite{ren2020fastspeech} to generate 944 hours of synthetic speech from Librispeech transcriptions, which is further used to generate synthetic video clips using LAM-LRS3-VSR-VL model. Using these 944 hours of synthetic data to train the BASE model under the LRS3 labeled data setting, the WER is improved from 36.7\% to 32.9\%, as shown in \cref{tab:6}. Although TTS-generated speech is not as natural as the original LibriSpeech data, the improvement of using synthetic data is still decent (3.8\%), indicating the further potential of SynthVSR in real-world applications where labeled video data is sparse, such as the healthcare industry.

\begin{table}[h]
\centering
\resizebox{0.4\textwidth}{!}{
\begin{tabular}{ccc}

\Xhline{2\arrayrulewidth}
\textbf{Training data} & \textbf{Hours} & \textbf{WER (\%)}                   \\ \hline
LRS3  & 438 & 36.7 \\
LRS3 + LBS-Synth & 1,382 & 30.8 \\
LRS3 + TTS-LBS-Synth & 1,382 & 32.9 \\
\Xhline{2\arrayrulewidth}
\end{tabular}}
\caption{Experiments results of using the synthetic speech generated synthetic video (TTS-LBS-Synth) for VSR training.}
\label{tab:6}
\vspace{-1em}
\end{table}
\vspace{-0.5em}
\section{Limitations \& Societal Impact}
The excellent performance improvement of SynthVSR comes at the cost of high computational demands during training, as we use additional large-scale synthetic video data. 
Also, although SynthVSR has achieved state-of-the-art VSR performance, the LRS3 test set is relatively limited (0.9 hours), which is from TED talks, while real-world videos are more challenging (e.g., egocentric videos). SynthVSR has many positive real-world applications, such as helping the hearing-impaired or people with aphonia with everyday communication. 

\section{Conclusion}
We have presented a semi-supervised method for VSR enhanced with synthetic lip movements. The speech-driven lip animation model is proposed to generate synthetic video data from labeled speech datasets and face images for scaling up VSR. Our method achieves state-of-the-art results on LRS3, outperforming prior work trained on more labeled or unlabeled real video data. Our work fosters future research on generating and exploiting synthetic visual data for VSR.

{\small
\bibliographystyle{ieee_fullname}
\bibliography{egbib}
}
\appendix
\section{Multimedia Video}
The multimedia video file (\url{https://www.youtube.com/watch?v=idIYKCzEFMU}) presents examples generated by the LAM-LRS3-AVoX-VSR model (as described in Sec. 4.3) with cropped lip images from CelebA and speech clips from Librispeech. We recommend turning on speakers to note the lip-syncing performance.

\section{Architecture Details}
\subsection{VSR Model}
\begin{figure}[t]
  \centering
   \includegraphics[width=0.85\linewidth]{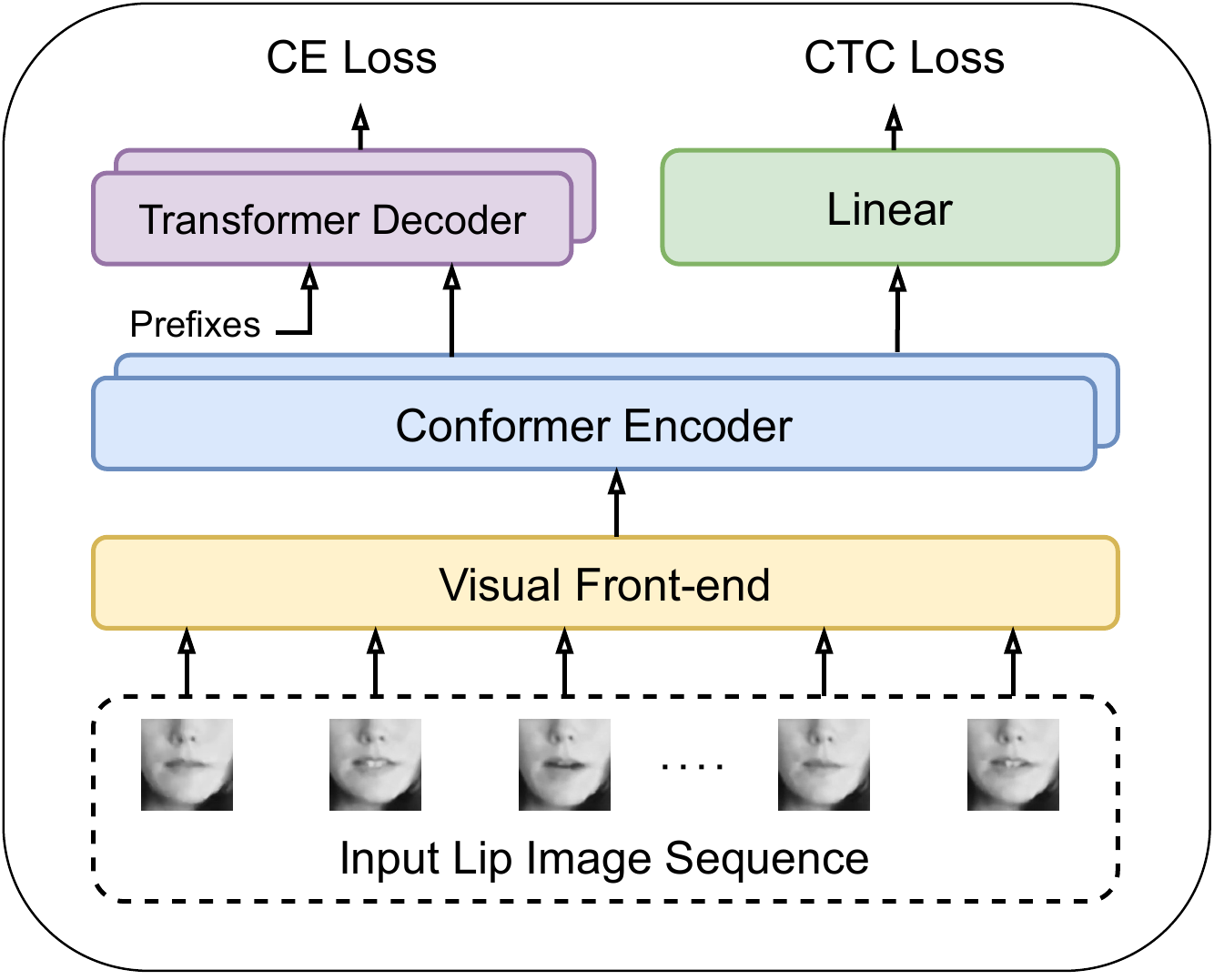}
   \caption{The VSR model used in this work is based on a Conformer encoder, a 3D ResNet visual front-end, and a combination of CTC and attention-based decoder.}
       \vspace{-1.5em}
   \label{fig:vsr}
\end{figure}

Here, we describe the details of the VSR model (as referenced in Sec. 3.1 of the main paper), which is the same as that used in the previous work \cite{ma2022nature, ma2022auto}. The architecture of the VSR model is depicted in \cref{fig:vsr}. The visual front-end consists of a 3D convolutional layer with a kernel size of $5 \times 7 \times 7$ followed by a ResNet-18 model. The visual features produced by the last residual block are aggregated along the spatial dimension by a global average pooling layer. Next, we use the Conformer encoder to model the visual features extracted by the front-end. Each Conformer block has a feed-forward module, a self-attention module, a convolution module, and a second feed-forward module stacked in order. We first use a linear layer to project the front-end embedding to a $D$-dimensional space, where $D$ is the dimension of the Conformer encoder input embedding. The projected features are added with the relative position information and further passed through the Conformer encoder backbone. Then, we use the Transformer decoder to map text the visual representation to a distribution over word-piece tokens. The decoder is composed of an embedding layer and a stack of Transformer decoder blocks, each decoder block consists of a self-attention module, an encoder-decoder cross-attention layer, and a feed-forward layer. Layer normalization is added  before each module. The prefixes from index 1 to $l$ - 1 are projected to embedding vectors, where $l$ is the length of target tokens. The absolute positional encoding is added to the embedding. The decoder generates the token sequence $y={(y_1, y_2, ..., y_l)}$ in autoregressive manner by factorising the joint probability distribution:
\begin{equation}
P_{Decoder}(y|z_{e}) = \prod_{i=1}^{l}P(y_i|z_{e}, y_{1:i-1}),
  \label{eq5}
\end{equation}

where $l$ is the length of the target token sequence and $z_e$ is the features extracted by the Conformer encoder.

We use a combination of CTC loss and attention-based CE loss as the training objectives of the baseline VSR model. A linear projection layer is used to map the high-level visual feature sequence into the output probabilities to compute the CTC loss. The CTC criterion assumes conditional independence between the output predictions and the estimated sequence posterior has the form of $P_{CTC}(y|x)\approx\prod_{t=1}^{T}p(y_t|x)$, and the CTC loss is defined as $\mathcal{L}_{CTC}=-\operatorname{log}P_{CTC}(y|x)$, where $x$ is the input video. The attention-based CE loss is calculated based on Equation (1): $\mathcal{L}_{CE}=-\operatorname{log}P_{Decoder}(y|z_{e})$. The VSR training objective is computed as follows:
\begin{equation}
\mathcal{L} = \alpha\mathcal{L}_{CTC} + (1-\alpha)\mathcal{L}_{CE},
  \label{eq5}
\end{equation}
where $\alpha$ controls the relative weight in CTC and CE losses. $\alpha$ was set to 0.1 in this work. In the evaluation, we use the model averaged over the last 10 checkpoints for decoding. 

\subsection{Speech-Driven Lip Animation}
We describe the implementation details (as referenced in Sec. 4.3 of the main paper) of the proposed speech-driven lip animation model. Specifically, the image encoder is a 5-layer 2D CNN. Batch normalization and ReLU activation are used for the first four layers while the last layer uses tangent activation. The image encoder maps a $96 \times 96$ input image to a 512-dimensional latent representation. The speech encoder is a stack of six 1D CNN followed by a 2-layer GRU. Batch normalization and ReLU activation are used for the first five layers while the last layer uses tangent activation. After that, the encoded speech chunks are fed to the GRU layers, which produce a 256-dimensional latent feature. We use StyleGAN2 as the frame decoder. Instead of generating frames from a constant input, our StyleGAN2 decoder uses the penultimate layer of the image encoder. The frame discriminator is a 5-layer CNN that determines whether a frame is real or not conditioned on the target frame. Batch normalization and Leaky ReLU activation are used after each convolution layer except for the last layer. The sequence discriminator is a 5-layer spatial-temporal CNNs, followed by a GRU layer and a single classifier layer. Batch normalization and ReLU activation are used first four layers and the fifth layer uses tangent activation. The detailed configurations of the image encoder, the speech encoder, the frame discriminator and the sequence discriminator are listed in \cref{tab:sm-1}, \cref{tab:sm-2}, \cref{tab:sm-3}, \cref{tab:sm-4}, respectively. We refer to the configuration of a convolutional layer as \textit{Conv[(kernel size), (stride), (padding) @ Channels]}, \textit{BN} and \textit{Tanh} indicates the Batch normalization and tangent activation, respectively.

\begin{table}[h]
\centering
\resizebox{0.4\textwidth}{!}{
\begin{tabular}{@{}cc@{}}
\toprule
\textbf{Layers} & \textbf{Image Encoder}                        \\ \midrule
1      & Conv2d{[}(4, 4) (2, 2) (1, 1) @ 64{]}, BN, ReLU  \\\hline
2      & Conv2d{[}(4, 4) (2, 2) (1, 1) @ 128{]}, BN, ReLU \\\hline
3      & Conv2d{[}(4, 4) (2, 2) (1, 1) @ 256{]}, BN, ReLU \\\hline
4      & Conv2d{[}(4, 4) (2, 2) (1, 1) @ 512{]}, BN, ReLU \\\hline
5      & Conv2d{[}(6, 6) (1, 1) (0, 0) @ 512{]}, Tanh     \\ \bottomrule
\end{tabular}}
\caption{Architecture of the image encoder.}
\label{tab:sm-1}
\vspace{-1.5em}
\end{table}

\begin{table}[h]
\centering
\resizebox{0.4\textwidth}{!}{
\begin{tabular}{@{}cc@{}}
\toprule
\textbf{Layers} & \textbf{Speech Encoder}                        \\ \midrule
1      & Conv1d{[}(80,) (16,) (32,) @ 16{]}, BN, ReLU  \\\hline
2      & Conv1d{[}(4,) (2,) (1,) @ 32{]}, BN, ReLU \\\hline
3      & Conv1d{[}(4,) (2,) (1,) @ 64{]}, BN, ReLU \\\hline
4     & Conv1d{[}(4,) (2,) (1,) @ 128{]}, BN, ReLU \\\hline
5     & Conv1d{[}(10,) (5,) (3,) @ 256{]}, BN, ReLU \\\hline
6      & Conv1d{[}(5,) (1,) (0,) @ 256{]}, Tanh     \\\hline
7      & GRU @ 256 \\\hline
8      & GRU @ 256 \\\bottomrule
\end{tabular}}
\caption{Architecture of the speech encoder.}
\label{tab:sm-2}
\vspace{-1.5em}
\end{table}

\begin{table}[h]
\centering
\resizebox{0.45\textwidth}{!}{
\begin{tabular}{@{}cc@{}}
\toprule
\textbf{Layers} & \textbf{Frame Discriminator}                        \\ \midrule
1      & Conv2d{[}(4, 4) (2, 2) (1, 1) @ 32{]}, BN, LeakyReLU  \\\hline
2      & Conv2d{[}(4, 4) (2, 2) (1, 1) @ 64{]}, BN, LeakyReLU \\\hline
3      & Conv2d{[}(4, 4) (2, 2) (1, 1) @ 128{]}, BN, LeakyReLU \\\hline
4      & Conv2d{[}(4, 4) (2, 2) (1, 1) @ 256{]}, BN, LeakyReLU \\\hline
5      & Conv2d{[}(6, 6) (1, 1) (0, 0) @ 1{]}     \\ \bottomrule
\end{tabular}}
\caption{Architecture of the frame discriminator.}
\label{tab:sm-3}
\vspace{-1.5em}
\end{table}

\begin{table}[h]
\centering
\resizebox{0.45\textwidth}{!}{
\begin{tabular}{@{}cc@{}}
\toprule
\textbf{Layers} & \textbf{Sequence Discriminator}                        \\ \midrule
1      & Conv3d{[}(7, 4, 4) (1, 2, 2) (3, 1, 1) @ 64{]}, BN, ReLU  \\\hline
2      & Conv3d{[}(1, 4, 4) (1, 2, 2) (0, 1, 1) @ 128{]}, BN, ReLU \\\hline
3      & Conv3d{[}(1, 4, 4) (1, 2, 2) (0, 1, 1) @ 256{]}, BN, ReLU \\\hline
4      & Conv3d{[}(1, 4, 4) (1, 2, 2) (0, 1, 1) @ 256{]}, BN, ReLU \\\hline
5      & Conv3d{[}(1, 6, 6) (1, 1, 1) (0, 0, 0) @ 128{]}, Tanh \\\hline
6      & GRU @ 512     \\\hline
7      & Linear @ 1 \\ \bottomrule
\end{tabular}}
\caption{Architecture of the sequence discriminator.}
\label{tab:sm-4}
\vspace{-1.5em}
\end{table}

\section{Experimental Details and Results}
\subsection{VSR Pre-training in Low-Resource Setting}
Here, we discuss the VSR pre-training details as referenced in Sec. 4.4 of the main paper. Because supervised training VSR models from scratch with long sequences often pose optimization problems, we first use 30 hours of LRS3 and 944 hours of Librispeech synthetic data to pre-train a SMALL VSR model with 12-layer Conformer encoder, 6-layer Transformer decoder, 256 input dimensions, 2048 feed-forward dimensions and 4 attention heads (encoder and decoder have same dimensions and attention heads). The SMALL model is further fine-tuned using 30 hours of LRS3 data. We pre-train and fine-tune the SMALL model for 75 and 25 epochs, respectively. The VSR WER after pre-training and fine-tuning are 58.9\% and 52.6\%, respectively, as shown in \cref{tab:sm-5}. The other training hyperparameters are the same as we used in Sec. 4.3. As the SMALL model has the same visual front-end as the BASE model, we initialize the visual front-end weights of the BASE model from the fine-tuned SMALL model for the low-resource labeled data setting.

\begin{table}[h]
\centering
\resizebox{0.45\textwidth}{!}{
\begin{tabular}{ccc}
\Xhline{2\arrayrulewidth}
\textbf{VSR model} & \textbf{Training data} & \textbf{WER} (\%)                  \\ \hline
SMALL-pretrain & LRS3 (30 hrs) + LBS-Synth  & 58.9 \\
SMALL-finetune & LRS3 (30 hrs) & 52.6 \\
\Xhline{2\arrayrulewidth}
\end{tabular}}
\caption{WER of pre-trained (SMALL-pretrain) and fine-tuned (SMALL-finetune) SMALL VSR models. WER is calculated without using the language model.}
\label{tab:sm-5}
\vspace{-1em}
\end{table}

\subsection{VSR Baseline in High-Resource Setting}
In Sec. 4.6 of the main paper, we only consider the BASE VSR model trained with 438 hours of LRS3 and 2630 hours of pseudo-labeled data as the baseline system, since we found LARGE model suffers from some convergence problems during training, which was potentially caused by its much larger model size.


\subsection{Experimental Results for Synthetic Video Data with Multiple Lip Images.}
Here, we studied the effect of the scale of Librispeech synthetic data (LBS-Synth) generated from the LAM-LRS3-VSR-VL model. We conducted one additional experiment using the BASE VSR model under the LRS3 labeled data setting. 
In Table~\ref{tab:Ablations_synth}, we show that double the size of LBS-Synth leads to further improvement (WER 30.8\% to 30.1\%). This is done by synthesizing two videos per speech clip with two CelebA lip images. This experiment further proves the flexibility and scalability of our synthetic data generation pipeline, which in principle could lead to unlimited video data for scaling up VSR. 

\begin{table}[h]
\centering
\resizebox{0.45\textwidth}{!}{
\begin{tabular}{ccc}
\Xhline{2\arrayrulewidth}
\textbf{Training data} & \textbf{Hours} & \textbf{WER w.o. LM (\%)}                   \\ \hline 
LRS3 + LBS-Synth x1 & 438 + 944 & 30.8 \\
LRS3 + LBS-Synth x2 & 438 + 1,888 & 30.1 \\

\Xhline{2\arrayrulewidth}
\end{tabular}}
\label{tab:5}
\caption{Multiple lip images coupled with a single speech leads to better performance. WER is calculated without using the language model.}
\vspace{-1em}
\label{tab:Ablations_synth}
\end{table}

\end{document}